\pdfoutput=1
\documentclass[11pt]{article}

\usepackage{ACL2023}

\usepackage{times}
\usepackage{latexsym}

\usepackage[T1]{fontenc}
\usepackage[utf8]{inputenc}

\usepackage{microtype}

\usepackage{inconsolata}

\usepackage{booktabs}
\usepackage{multirow}
\usepackage{microtype}
\usepackage{amsmath}
\usepackage{amssymb}
\usepackage{amsfonts}
\usepackage{bbm}
\usepackage{diagbox}
\usepackage{graphicx}
\usepackage{makecell}
\usepackage{xspace}
\usepackage[normalem]{ulem}
\usepackage{mathabx}
\usepackage{xcolor}
\usepackage[normalem]{ulem}
\newcommand\hly{\bgroup\markoverwith
    {\color{yellow}{\rule[-.5ex]{.1pt}{2.5ex}}}\ULon}

\newcommand{\biasamplified}[0]{\textit{bias-amplified}\xspace}

\newcommand{\easy}[0]{\textit{biased}\xspace}
\newcommand{\hard}[0]{\textit{anti-biased}\xspace}
\newcommand{\Easy}[0]{\textit{Biased}\xspace}
\newcommand{\Hard}[0]{\textit{Anti-biased}\xspace}

\newcommand{\cartography}[0]{\textsl{dataset cartography}\xspace}

\newcommand{\minority}[0]{\textsl{minority examples}\xspace}

\newcommand{\partialinput}[0]{\textsl{partial-input}\xspace}

\newcommand{\tabref}[1]{Tab.~\ref{tab:#1}}
\newcommand{\figref}[1]{Fig.~\ref{fig:#1}}
\newcommand{\appref}[1]{App.~\ref{#1}}
\newcommand{\secref}[1]{Sec.~\ref{sec:#1}}
\newcommand{\subsecref}[1]{Sec.~\ref{subsec:#1}}

\newcommand{\base}{\textsc{base}\xspace}
\newcommand{\modellarge}{\textsc{large}\xspace}
\newcommand{\roberta}{\textsc{RoBERTa}\xspace}
\newcommand{\deberta}{\textsc{DeBERTa}\xspace}

\newcommand{\qqp}{{\textsl{QQP}}\xspace}
\newcommand{\mnli}{{\textsl{MultiNLI}}\xspace}
\newcommand{\anli}{\textsl{{ANLI}}\xspace}

\newcommand{\hans}{\textsl{{HANS}}\xspace}
\newcommand{\paws}{\textsl{{PAWS}}\xspace}

\newcommand{\wanli}{\textsc{{WaNLI}}\xspace}
\newcommand{\sst}{\textsl{{SST-2}}\xspace}

\newcommand{\deepcluster}{${{\textsc{DeepCluster}}}$\xspace}

\newcommand{\datatrain}{$\mathcal{D}^\text{train}$\xspace}

\newcommand{\datatraineasy}{$\mathcal{D}^\text{train}_\text{\textsl{biased}}$\xspace}

\newcommand{\datatesthard}{$\mathcal{D}^\text{test}_\text{\textsl{anti-biased}}$\xspace}

\definecolor{mustard}{rgb}{1.0, 0.86, 0.35}
\title{Fighting Bias with Bias:\\
Promoting Model Robustness by Amplifying Dataset Biases}

\author{Yuval Reif \quad\quad Roy Schwartz\\
  School of Computer Science and Engineering, The Hebrew University of Jerusalem, Israel \\
  \texttt{\{yuval.reif,roy.schwartz1\}@mail.huji.ac.il}}

\begin{document}
\maketitle

\begin{abstract}
NLP models often rely on superficial cues known as \emph{dataset biases} to achieve impressive performance, and can fail on examples where these biases do not hold. 
Recent work sought to develop robust, unbiased models by filtering \easy examples from training sets.  
In this work, we argue that such filtering can obscure the true capabilities of models to overcome biases, which might never be removed in full from the dataset.
We suggest that in order to drive the development of models robust to subtle biases, dataset biases should be \emph{amplified} in the training set. 
We introduce an evaluation framework defined by a \biasamplified training set and an \hard test set, both automatically extracted from existing datasets.
Experiments across three notions of \emph{bias}, four datasets and two models show that our framework is 
substantially more challenging for models than the original data splits, and even more challenging than hand-crafted challenge sets. 
Our evaluation framework can use any existing dataset, even those considered obsolete, to test model robustness. 
We hope our work will guide the development of robust models that do not rely on superficial biases and correlations.
To this end, we publicly release our code and data.\footnote{
\url{https://github.com/schwartz-lab-NLP/fight-bias-with-bias}
}

\end{abstract}
\section{Introduction}
\begin{figure}[t]
\centering
\includegraphics[width=\columnwidth]{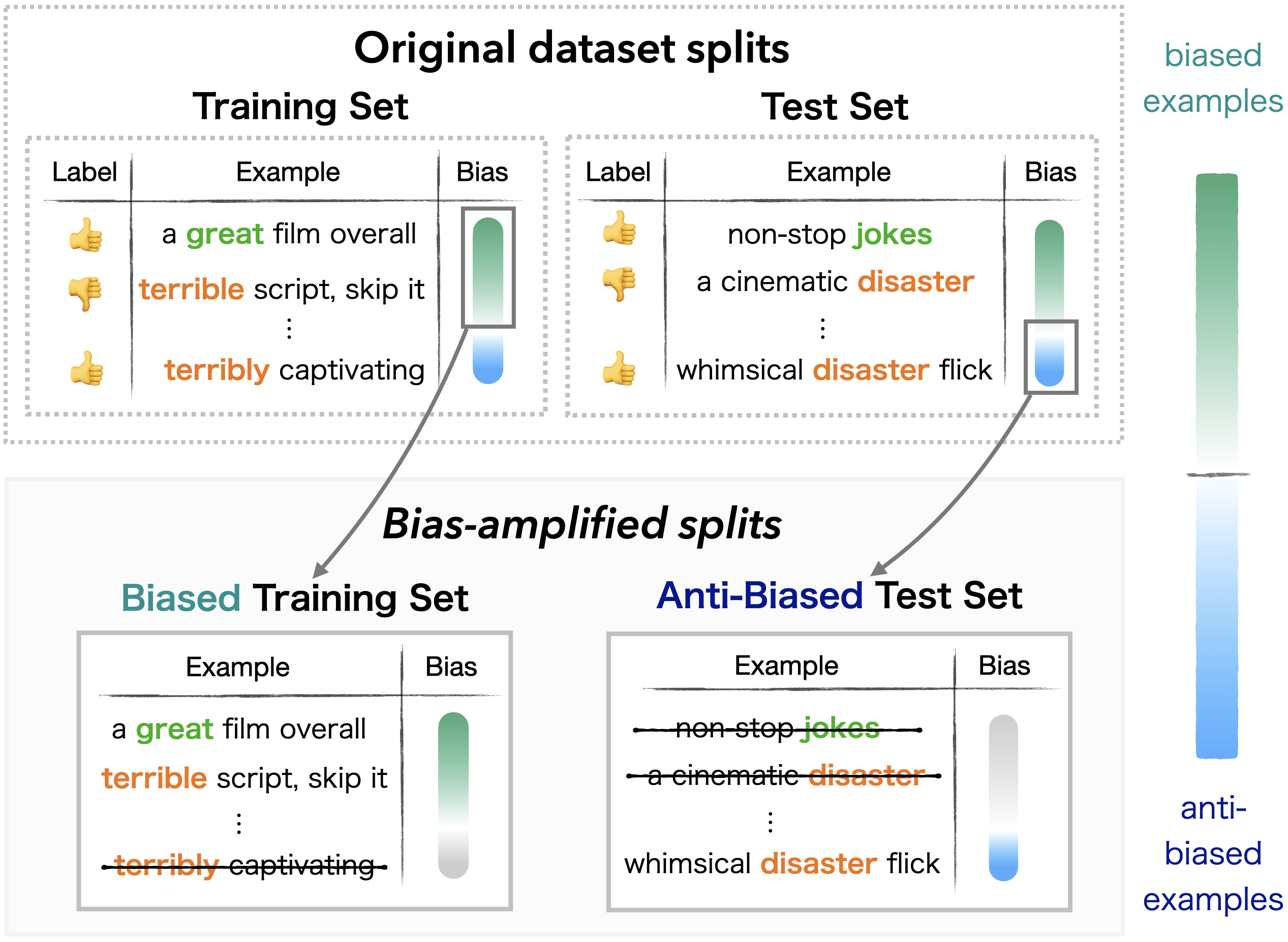}
\caption{ 
To guide the development of models robust to subtle biases, we propose to extract bias-amplified splits for existing benchmarks.
Our approach first partitions a given dataset into \easy and \hard instances.
It then constructs a \easy training set and an \hard test set, which are used to evaluate model generalization.
}
\label{fig:method}
\end{figure}
 
NLP models often exploit repetitive patterns introduced during data collection, known as \emph{dataset biases}, to achieve strong performance~\cite{poliak-etal-2018-hypothesis,mccoy-etal-2019-right}.\footnote{Instances that can be solved using such biases are typically referred to as ``\easy''~\cite{he-etal-2019-unlearn}.}
This trend has led to attempts of improving the evaluation of NLP models by creating test sets that are different from the training sets, e.g., from a different domain~\cite{williams-etal-2018-broad} or a different distribution~\cite{wilds}, and challenge sets that focus on counterexamples to known biases in the training set, which we refer to as \hard examples~\cite{jia-liang-2017-adversarial, naik-etal-2018-stress,utama-etal-2020-towards}. 

To address these gaps, some works used balancing techniques to create \textit{unbiased} datasets, by filtering out \easy examples~\cite{zellers-etal-2018-swag,aflite, swayamdipta-etal-2020-dataset}, or injecting \hard examples into the training sets~\cite{nie-etal-2020-adversarial,wanli}.
In this work we argue that in order to encourage the development of robust models, we should in fact \textbf{amplify} biases in the training sets, while 
adopting the challenge set approach and 
making test sets \hard (\figref{method}). 

\begin{figure}[t]
\centering
\includegraphics[width=\columnwidth]{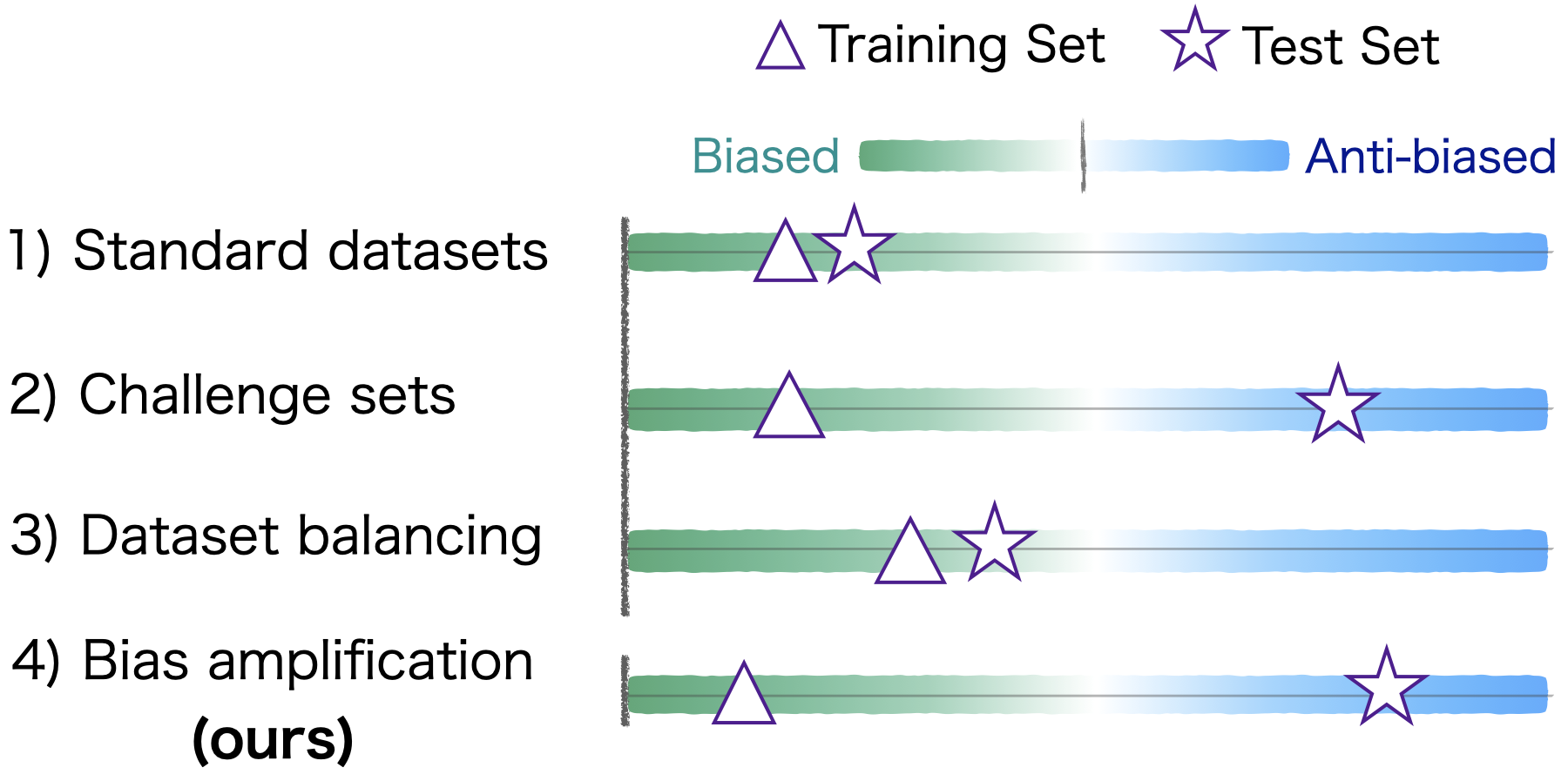}
\caption{ 
Different approaches to data collection. 
In standard datasets (1), the training and test sets mostly contain a majority of \easy instances. Challenge sets (2) curate \hard test sets. Balancing and filtering methods (e.g., adversarial filtering, 3) collect \textit{unbiased} training and test sets. Our framework (4) contains \easy training sets and \hard test sets.
}
\label{fig:approaches}
\end{figure}

Amplifying dataset biases might seem counter-intuitive at first. Our work follows recent work that challenged the assumption that biases can ever be fully removed from a given dataset~\cite{schwartz-stanovsky-2022-limitations}, arguing that models are able to pick up on very subtle phenomena even in partially balanced (or mostly unbiased) datasets~\cite{gardner-etal-2021-competency}.
As a result, dataset balancing, while potentially improving generalization, might make it harder to develop models that are resilient to such biases; these biases ``hide'' in the balanced training sets, and the way models handle them is hard to evaluate and make progress on.\footnote{Indeed, training on adversarial data doesn't necessarily generalize to non-adversarial data~\cite{kaushik-etal-2021-efficacy}.} Instead, we argue that academic benchmarks should include training splits that mainly consist of \easy examples (see \figref{approaches}). Such splits will drive the development of robust models that generalize beyond biases, ideally even subtle ones.

We present a simple method to implement our approach (\secref{framework}). 
Given a dataset in which both training and test sets are divided into \easy and \hard subsets, we remove the \hard instances from the training set and the \easy ones from the test set. The new splits then form a challenging evaluation setting.
We assume that \easy instances constitute the majority of a dataset~\cite{gururangan-etal-2018-annotation,utama-etal-2020-towards}, and thus the resulting training sets are similar in size to the original ones (though the test sets are smaller).

To discern \easy and \hard instances, we consider three model-based approaches (\secref{methods}): (a) dataset cartography~\cite{swayamdipta-etal-2020-dataset}, 
which uses training dynamics to profile the difficulty of learning individual data instances.
In this approach, we identify instances that are hard-to-learn as \hard~\cite{sanh2021learning, he-etal-2019-unlearn}; (b) partial-input models~\cite{kaushik-lipton-2018-much, poliak-etal-2018-hypothesis}, which 
are forced to rely on bias, regarding instances on which they fail as \hard; and a method we introduce for identifying (c) \emph{minority examples}~\cite{tu-etal-2020-empirical, sagawa2020investigation}, 
which groups a dataset's instances using deep clustering~\cite{deepcluster} and regards the minority-label instances within each cluster as \hard.

We apply our framework to \mnli~\cite{williams-etal-2018-broad} and \qqp~\cite{wang-etal-2018-glue}, on which trained models exceed human performance.
We also experiment with two datasets that are considered more challenging: Adversarial NLI (\anli;~\citealp{nie-etal-2020-adversarial}) and \wanli~\cite{liu-etal-2022-wanli}.
We use a \roberta-\base~\cite{roberta} model for selecting \easy and \hard instances according to each method, and evaluate the performance of \roberta and \deberta~\cite{deberta} \modellarge models under our proposed setting (\secref{results}).
While \hard instances are naturally challenging for models, amplifying biases in the training set makes them even more challenging; using the partial-input and minority examples methods, we observe mean absolute performance reductions of $15.8\%$ and $31.8\%$, respectively. Using instances detected with dataset cartography leads to smaller (though still large) reductions of $10.1\%$.

We compare \textsl{bias-amplified splits} to hand-crafted challenge sets such as \hans~\cite{mccoy-etal-2019-right}, 
and find that our automatically-generated \hard test sets are both of similar difficulty to such challenge sets, and capture a more diverse set of biases. 
Our framework can further be used to augment existing challenge sets, as training on bias-amplified data increases their difficulty.

Next, we investigate how many \hard examples are required for generalization, by gradually re-inserting such instances to the training set~\cite{liu-etal-2019-inoculation}. 
While models greatly benefit from observing small amounts of \hard instances, 
\hard test sets remain challenging, 
and additional performance gains require much larger quantities (\secref{reinsert}). 
We then show that standard debiasing methods applied to bias-amplified training sets lead to little to no gains in performance (\secref{mitigating}).

Our findings may change the way we evaluate the robustness of NLP models, and in particular their level of generalization beyond the biases of their training sets. Our method requires no new annotation or any task-specific expertise. It allows to rejuvenate datasets previously considered as obsolete, and thus reuse the intensive efforts used in their curation. 
We release our new dataset splits along with code for automatically creating bias-amplified splits for other datasets.

\section{Amplifying Dataset Biases to Advance Model Robustness}
\label{sec:framework}

This section motivates our approach in view of recent developments in NLP, provides a general overview of the framework we use to implement it, and discusses its applications.

\subsection{Motivation: Data Balancing Hides Biases}
This paper focuses on the problem of creating robust models that generalize beyond dataset biases. 
A common approach to addressing this problem is removing these biases from the training data~\cite{zellers-etal-2018-swag,aflite}. This approach is intuitive---if a model doesn't observe these biases in the first place, it is less likely to learn them, and will thus generalize better.

Despite the appeal of this approach, it suffers from several problems. First, recent work has argued that models are sensitive to very fine-grained biases, which are hard to detect and filter~\cite{gardner-etal-2021-competency}. Other works have shown that training on bias-filtered datasets does not necessarily lead to better generalization~\cite{kaushik-etal-2021-efficacy, parrish-etal-2021-putting-linguist}, indicating that while such training sets are less biased, models might still rely on biases to solve them. Finally, recent studies argued that even with our utmost efforts, we may never be able to create datasets that contain no exploitable biases~\cite{linzen-2020-accelerate,schwartz-stanovsky-2022-limitations}.

As a result, this paper argues that mitigating the negative effect of dataset biases is not only a data problem, but needs to also come from better \textit{modeling}. 
But how can we create a testbed for developing models that overcome these biases? We argue that training on datasets filtered for such biases will not suffice in developing such models, and in fact make it harder to do so; 
as subtle biases still ``hide'' inside filtered training sets, it is much harder to track them, evaluate their impact and importantly---develop models that learn to ignore them.

Instead, in this paper we propose that when evaluating model robustness, dataset biases should be \textbf{amplified} 
by training mostly on \easy instances, while using \hard instances for evaluation (\figref{approaches}).
This simple setting defines a challenging test, where models must counteract dataset biases and learn generalizable solutions in order to succeed, as the \hard test set cannot be solved using the \easy training set's statistical cues.

\subsection{Framework for Amplifying Dataset Biases}

We describe our approach for amplifying dataset biases during training to evaluate model generalization. Given a dataset split into training and test sets $\mathcal{D} = \mathcal{D}^\text{train} \cup \mathcal{D}^\text{test}$, we begin by dividing its instances across both splits into \easy and \hard subsets.\footnote{We consider three different notions of \easy instances (\secref{methods}), but other definitions (e.g.,~\citealp{godbole-jia-2023-benchmarking}, see \secref{related_work} for discussion) could be integrated into our framework.}
To evaluate a model's robustness, we first train it on the portion of \easy train instances \datatraineasy. We assume most data instances are \easy~\cite{gururangan-etal-2018-annotation}, so this process results in small reductions in training set sizes compared to \datatrain.
We then evaluate the model on the \hard test instances \datatesthard, and compare it to the performance of the same model trained on the full training set. Drops in performance between the two indicate that the model struggles to overcome its training set biases.

\subsection{Discussion}

\paragraph{Applications}
We suggest our framework as a tool for studying and evaluating models. As such, it is orthogonal to data collection procedures. Importantly, we do not suggest to intentionally collect biased data when curating new datasets.
Nonetheless, data collected in large quantities tends to contain unintended regularities~\cite{gururangan-etal-2018-annotation}. 
We therefore propose to use bias-amplified splits to complement benchmarks with challenging evaluation settings that test model robustness, in addition to the dataset's main training and test sets.

Such splits, when created using the methods we consider in this work, can be created automatically and efficiently for any dataset.
These include newly collected datasets, but also existing ones, such as obsolete benchmarks on which model performance is too high to measure further progress, allowing for the rejuvenation and reuse of benchmarks.

\paragraph{\Hard vs.~challenge sets}
Our framework provides an evaluation environment to assess model robustness, similar to challenge sets. 
However, unlike challenge sets, which are often manually curated with protocols designed to create difficult examples, our approach is automatic and uses data collected using the exact same protocol as the model's training data. 
Still, we find that \hard test sets are challenging for models and can capture more diverse biases, and moreover---that training on \biasamplified data further enhances the difficulty of existing challenge sets (\subsecref{main_results}).
Consequently, our framework can be employed to evaluate robustness in tasks where challenge sets are unavailable, or in conjunction with existing challenge sets for a more comprehensive evaluation.

\paragraph{Can models generalize from \easy data?}
A natural question to ask about our approach is whether we can truly expect models to generalize from a biased training distribution. Although the \easy training sets could be solved by capturing only a subset of relevant features, their instances can still provide valuable information for learning additional features that are important for generalization yet under-utilized by models~\cite{simplicitybias, geirhos2020shortcut}. 
Previous work has proposed techniques to encourage models to learn diverse, unbiased representations from extremely biased training distributions, mostly focusing on domains outside of NLP~\cite{kim2019learning, bahng2020learning,gradientstarvation}. This is likely due to the difficulty of defining and controlling biased distributions in textual domains. 
Our work paves the way for implementing and evaluating such methods specifically for NLP.

Related to this concern is our decision to leave \textbf{no} \hard instances in the training set. 
Indeed, it is likely that for many biases, at least \textit{some} counter-examples will be found in the training set.
We admit that this decision is not a major component of our approach, and it could be easily implemented with a \textit{small} number of \hard instances in the training set instead.
To avoid deciding on the numeric definition of small, and to make the setup as challenging as possible, we experiment throughout this paper with no (identified) \hard instances in training.
In \secref{reinsert} we study the effect of using limited amounts of such counter-examples, by reinserting some \hard instances into training.

\section{Definitions of \Easy and \Hard Examples}
\label{sec:methods}

Our approach requires a drop-in method for classifying a dataset's examples into \easy and \hard instances.
We consider the following model-based methods for doing so. We note that none of them requires any prior knowledge or task-specific expertise. All methods can be computed automatically at the reasonable cost of training and evaluating a (possibly smaller) model on the dataset.

\paragraph{Dataset Cartography}~\cite{swayamdipta-etal-2020-dataset}
is a method to automatically characterize a dataset's instances according to their contribution to a model's performance, by tracking a model's training dynamics. Specifically, measuring each instance's \textbf{confidence}---the mean of the predicted model probabilities for the gold label across training epochs---reveals a region of \textit{easy-to-learn} instances with high confidence which the model consistently predicts correctly throughout training, and a region of low-confidence \textit{hard-to-learn} instances, on which the model consistently fails during training.
We follow previous work which considered instances that models find easy or hard to solve as more likely to be \easy or \hard, respectively~\cite{sanh2021learning, he-etal-2019-unlearn}.

To estimate the confidence of test instances, we make predictions with a partially trained model at the end of each epoch on the test set (as typically done on the validation set), and use the average confidence scores across epochs.\footnote{We emphasize that we are \textbf{not} fine-tuning on the test set, nor using it to select any hyperparameters. This process only annotates the \easy and \hard portions of the test set.} To choose \hard examples, we use the $q\%$ most \textit{hard-to-learn} instances in each of the training and the test sets individually, where $q$ is a hyperparameter. We consider all other examples as \easy.

\paragraph{Partial-input baselines}
is a common method for identifying annotation artifacts in a dataset. The method works by examining the performance of models that are restricted to using only part of the input. Such models, if successful, are bound to rely on unintended or spurious patterns in the dataset. Examples include question-only models for visual question answering~\cite{goyal2017making}, ending-only models for story completion~\cite{schwartz-etal-2017-effect} and hypothesis-only models for natural language inference~\cite{poliak-etal-2018-hypothesis}.

Held-out instances where such baselines fail are considered \hard and less likely to contain artifacts~\cite{gururangan-etal-2018-annotation}.\footnote{Such instances might still contain more complex artifacts that are only detectable when jointly inspecting all parts of the input~\cite{feng-etal-2019-misleading}.}
Generating a \easy training set for this method is not trivial, as the partial-input model is likely to fit to the training data during training, and thus almost all examples will be labeled \easy. We therefore follow the dataset cartography approach with a partial-input baseline, and  compute the mean confidence score for each instance across epochs. We  select the $q\%$ most hard-to-learn instances as \hard.

\paragraph{Minority examples}
Current models are typically sensitive to \textsl{minority examples} that defy common statistical patterns found in the rest of the data, especially when the amount of such examples in the training set is scarce~\cite{tu-etal-2020-empirical, sagawa2020investigation}.
Minority examples are often detected by heuristically searching for spurious features correlated with one label in the instances of another label (e.g., high word overlap between two non-paraphrase texts).
Motivated by recent work that leverages instance similarity in the representation space of fine-tuned language models for various use cases~\cite{liu-etal-2022-wanli, pezeshkpour-etal-2022-combining}, we propose a  model-based clustering approach to automatically detect minority examples. 

We follow a three-step approach. 
First, we cluster the \emph{training} set using [CLS] token representations extracted from a model trained on the dataset. 
Second, to detect minority examples in the training set, we inspect the distribution of instances over the task labels $L$ within each cluster $c_i$. We define a cluster $c_i$'s \textsl{majority label} as the label $\ell_{i}\in L$ associated with the most instances in $c_i$. We consider all other labels as $c_i$'s minority labels. Instances belonging to their cluster's minority labels are regarded as \textsl{minority examples}, and accordingly \hard, while and all others are considered \easy. 
Finally, to detect minority examples in the \emph{test} set, we extract [CLS] representations for all test instances, and assign each instance to the cluster of its nearest neighbor in the \emph{training} set using Euclidean distance. If the test instance $(x,y)$ is assigned to cluster $c_i$, we consider $(x,y)$ as a majority example iff it belongs to $c_i$'s majority label, i.e., if $y== \ell_{i}$.
\footnote{Note that unlike the methods described above for detecting \hard subsets, the minority examples approach does not require a pre-determined size for the resulting subset, as it is induced by the clustering.}

Our preliminary experiments show that standard clustering algorithms tend to create label-homogeneous clusters, i.e., they are less likely to cluster together instances from different labels. 
We thus use \deepcluster~\cite{deepcluster}, which we find to create more label-diverse clusters. \deepcluster alternates between grouping a model's representations with a standard clustering algorithm\footnote{We use Ward's method~\cite{ward1963}, a popular deterministic algorithm for hierarchical clustering which has the same objective function as K-means.} to produce pseudo-labels, and fine-tuning a new pretrained model to predict these pseudo-labels. 
We perform one iteration of deep clustering and then cluster the representations of the \deepcluster model to obtain the final clustering.
\appref{sec:app_clustering} shows details and preliminary results on alternative clustering methods.

\section{Models Struggle with Amplified Biases}
\label{sec:results}

\begin{table*}[t]
% \footnotesize
\centering
\resizebox{\textwidth}{!}{%
\setlength{\tabcolsep}{2.5pt}
\begin{tabular}{@{}llllllllllllllllll@{}}
\toprule
& \multicolumn{1}{c}{} &
  \multicolumn{4}{c}{\mnli} &
  \multicolumn{4}{c}{\qqp} &
  \multicolumn{4}{c}{\wanli} &
  \multicolumn{4}{c}{\anli} \\
& \multicolumn{1}{c}{} &
  \multicolumn{1}{c}{Orig.} &
  \multicolumn{1}{c}{Cart.} &
  \multicolumn{1}{c}{ParIn} &
  \multicolumn{1}{c|}{Mino.} &
  \multicolumn{1}{c}{Orig.} &
  \multicolumn{1}{c}{Cart.} &
  \multicolumn{1}{c}{ParIn} &
  \multicolumn{1}{c|}{Mino.} &
  \multicolumn{1}{c}{Orig.} &
  \multicolumn{1}{c}{Cart.} &
  \multicolumn{1}{c}{ParIn} &
  \multicolumn{1}{c|}{Mino.} &
  \multicolumn{1}{c}{Orig.} &
  \multicolumn{1}{c}{Cart.} &
  \multicolumn{1}{c}{ParIn} &
  \multicolumn{1}{c}{Mino.} \\ \midrule
   \parbox[t]{2mm}{\multirow{3}{*}{\rotatebox[origin=c]{90}{\textit{Train}}}} & \textit{full} &
  $90.4_{0.2}$ &
  $59.9_{0.7}$ &
  $79.7_{0.6}$ &
  \multicolumn{1}{l|}{$71.9_{0.3}$} &
  $92.0_{0.1}$ &
  $59.4_{0.4}$ &
  $78.6_{0.2}$ &
  \multicolumn{1}{l|}{$73.5_{0.3}$} &
  $76.2_{0.1}$ &
  $19.7_{3.4}$ &
  $59.5_{0.6}$ &
  \multicolumn{1}{l|}{$60.2_{1.6}$} &
  $55.4_{0.1}$ &
  $14.8_{0.6}$ &
  $34.9_{1.0}$ &
  $44.3_{0.4}$ \\
& \textit{rand} &
  $90.3_{0.1}$ &
  $59.5_{0.7}$ &
  $79.7_{0.7}$ &
  \multicolumn{1}{l|}{$71.9_{1.0}$} &
  $91.6_{0.1}$ &
  $57.8_{0.4}$ &
  $78.0_{0.5}$ &
  \multicolumn{1}{l|}{$71.8_{1.0}$} &
  $76.0_{0.1}$ &
  $17.6_{0.9}$ &
  $58.0_{3.0}$ &
  \multicolumn{1}{l|}{$59.2_{2.2}$} &
  $55.1_{0.7}$ &
  $15.7_{1.0}$ &
  $34.1_{1.4}$ &
  $44.2_{1.1}$ \\ \cmidrule(l){2-18}
& \textit{bias} &
  $88.4^{\bigast}_{0.7}$ &
  $51.7_{0.5}$ &
  $68.2_{0.3}$ &
  \multicolumn{1}{l|}{$50.5_{1.2}$} &
  $88.3^{\bigast}_{1.9}$ &
  $49.0_{0.3}$ &
  $60.3_{1.8}$ &
  \multicolumn{1}{l|}{$31.3_{0.4}$} &
  $74.9^{\bigast}_{1.0}$ &
  $13.7_{0.8}$ &
  $43.5_{2.9}$ &
  \multicolumn{1}{l|}{$25.8_{2.4}$} &
  $51.2^{\bigast}_{1.9}$ &
  $\phantom{0}5.8_{0.7}$ &
  $16.0_{1.0}$ &
  $12.3_{0.8}$ \\ \bottomrule
\end{tabular}

}
\caption{Accuracy of our approach with \roberta-\modellarge models. Different rows correspond to different training schemes: the full dataset (\textit{full}), a biased subset (\textit{bias}) and a random subset the size of \textit{bias} (\textit{rand}). Column groups correspond to different datasets. Individual columns represent testing schemes: the original validation/test set (Orig.)
and the \hard test splits: \cartography (Cart.), \partialinput (ParIn) and \minority (Mino.). Reported values are averaged across three random seeds, with standard deviation as subscripts.
Results in the last row (\textit{bias}) are of training on the \easy split and testing on the respective \hard split, except for Orig.~values (marked with $\bigast$), which are averaged over runs on all three \easy splits.
Model evaluation on bias-amplified splits results in weak performance on \hard test instances compared to the original data splits. 
} 
\label{tab:main_results}
\end{table*}

We next use our framework to evaluate the extent to which models generalize beyond the biases of their training sets.

\subsection{Experimental Setup}
\label{sec:exp_setting}
We create bias-amplified splits for four datasets: two 
(\qqp,~\citealp{wang-etal-2018-glue}; and \mnli, ~\citealp{williams-etal-2018-broad}) that were shown to contain considerable biases~\cite{zhang-etal-2019-paws, gururangan-etal-2018-annotation}; and two additional datasets (\anli,~\citealp{nie-etal-2020-adversarial}; and \wanli,~\citealp{liu-etal-2022-wanli}) designed to contain smaller proportions of \easy instances. QQP is a duplicate question identification dataset, while the other three are natural language inference (NLI) datasets.

We split all datasets into \easy and \hard parts according to each of the three methods described in \secref{methods}. We use a \roberta-\base~\cite{roberta} model for all three methods: we fine-tune the model on each dataset to compute training dynamics for \cartography, and also to extract and cluster [CLS] representations for identifying \minority; we separately train the model on partial inputs to obtain training dynamics for \textsl{partial-input baselines}. We use hypothesis-only baselines for NLI datasets. For QQP, we use the first question of each pair.

We then evaluate the performance of \roberta and \deberta ~\cite{deberta} \modellarge models under our proposed framework. We train models on the \easy training split obtained from each of the three methods, and report their performance on the corresponding \hard test sets.\footnote{We use the validation sets of \qqp and \wanli, and the validation-matched set for \mnli, as their test sets are not publicly available.} 
Since the number of \easy training instances is induced by the clustering in the \textsl{minority examples} approach, but is a hyperparameter $q$ for the two other approaches, we adjust $q$ to create equally sized training sets for all three methods. This results in $79\%$ of the training set for \mnli, $82\%$ for \qqp and \anli, and $87\%$ for \wanli.\footnote{When selecting \minority for \mnli and \qqp, we consider all labels but a cluster's \textsl{majority label} as minority labels. Using this setting for \anli and \wanli results in specifying more than $40\%$ of the training set as \minority. This leaves too few \easy instances for training and substantially changes the original training distribution. Therefore, for these datasets, we use the label with the \emph{least} instances within a cluster as its minority-label.} 
See \appref{sec:settings} for more details on the experimental setup.

\paragraph{Baselines}
We compare against two baselines: the original training split (\textbf{100\% train}) and a random sample of the same size as the \easy training splits (\textbf{random}). 
In addition to the \hard test set, we also report performance on the original test set to validate that the model's training data (the \easy training instances) is sufficient for learning the task.

\paragraph{Hyperparameters selection}
Our approach for identifying \minority is based on clustering the representations of a fine-tuned model. The clustering algorithm we use, \deepcluster (\secref{methods}), has three hyperparameters: the number of final clusters $k$, the number of pseudo-labels $m$ for representation learning, and the Transformer layer from which [CLS] representations are extracted for clustering. We use $k=10$ clusters for all datasets, and search for a good configuration for the other two hyperparameters on \sst~\cite{socher-etal-2013-recursive}: 
for each set of hyperparameters, we apply the \minority method to create \easy training and \hard test splits, and train two \roberta-\base models---one on the \easy training split, and a baseline model on an equally-sized random training subset. We select the hyperparameters that lead to the largest performance drop on \hard test instances between the two, 
and use them in all further experiments to cluster other datasets; 
see \appref{subsec:clustering_hyperparameters} for  details.

\subsection{Results}
\label{subsec:main_results}

\paragraph{Models struggle with \easy training sets}
\tabref{main_results} shows our results for \roberta-\modellarge. 
We observe that the baseline models struggle with all \hard test sets, even when training on the full training set. The \hard test splits based on \cartography prove to be the most initially difficult, with the splits created using the two other methods overall similar in difficulty.
Still, model performance on \hard instances drops further when training on \easy training splits; taking the mean across datasets, performance drops by 8.4\% for \cartography-based splits, 16.2\% for \partialinput, and 32.5\%  for \minority. 
Results for \deberta-\modellarge (\appref{deberta_main_results}) follow the same trends, with mean performance reductions of 11.8\% for \cartography, 15.4\% for \partialinput, and 31.1\%  for \minority. 

We also observe that training on \easy splits leads to minor reductions on the full test sets, indicating that while current models trained on our training splits fail to generalize beyond the biases in these sets, they are seemingly able to learn the tasks at hand.

\begin{table}[t]
% \footnotesize
\centering
\resizebox{\columnwidth}{!}{%

\begin{tabular}{@{}lcccc|ccc@{}}
\toprule
       & \multicolumn{4}{c|}{\mnli-\hard}                                           & \multicolumn{3}{c}{\hans}                                        \\
  & E            & N            & C                                 & Mean         & E            & $\neg$E                             & Mean         \\ \midrule
\emph{full} & $51.4_{3.5}$ & $75.5_{1.0}$ & \multicolumn{1}{c|}{$77.8_{2.3}$} & $71.9_{0.3}$ & $99.8_{0.2}$ & \multicolumn{1}{c|}{$56.5_{0.8}$}  & $78.2_{0.5}$ \\
\emph{rand} & $51.1_{3.2}$ & $75.6_{2.8}$ & \multicolumn{1}{c|}{$78.0_{1.6}$} & $71.9_{1.0}$ & $99.8_{0.1}$ & \multicolumn{1}{c|}{$55.7_{1.1}$}  & $77.8_{0.5}$ \\
\midrule
\easy  & $45.1_{1.7}$ & $48.9_{2.3}$ & \multicolumn{1}{c|}{$57.6_{1.4}$} & $50.5_{1.2}$ &  $99.8_{0.1}$ & \multicolumn{1}{c|}{$2.9_{0.9}$} & $51.4_{0.4}$ \\ \bottomrule
\end{tabular}

}

\caption{Accuracy of \roberta-\modellarge models trained on different subsets of \mnli, when evaluated on the dataset's \hard test split and on \hans.
Model performance is reported per label (Entailment: \emph{E}, Neutral: \emph{N}, Contradiction: \emph{C}, Not entailment: \emph{$\neg$E}) and over all examples (\emph{Mean}). \Easy and \hard splits are created with the \minority method. 
\Hard sets are comparably difficult to manually designed challenge sets, yet capture diverse biases in \emph{all} task labels.
}
\label{tab:hans}
\end{table}

\begin{table}[t]
\centering
\footnotesize
\resizebox{\columnwidth}{!}{%

\begin{tabular}{@{}lccc|ccc@{}}

\toprule
       & \multicolumn{3}{c|}{\qqp-\hard}                                 & \multicolumn{3}{c}{\paws}                                       \\
  & D            & $\neg$ D                            & Mean         & D            & $\neg$ D                            & Mean         \\ \midrule
\emph{full} & $80.9_{0.4}$ & \multicolumn{1}{c|}{$69.9_{0.5}$} & $73.5_{0.3}$ & $94.2_{0.6}$ & \multicolumn{1}{c|}{$17.7_{3.5}$} & $51.5_{1.7}$ \\
\emph{rand} & $80.0_{0.4}$ & \multicolumn{1}{c|}{$67.8_{1.3}$} & $71.8_{1.0}$ & $95.2_{0.3}$ & \multicolumn{1}{c|}{$13.4_{1.5}$} & $49.6_{0.9}$ \\
\midrule
\easy  & $27.9_{2.3}$ & \multicolumn{1}{c|}{$33.0_{1.4}$} & $31.3_{0.4}$ & $95.6_{0.8}$ & \multicolumn{1}{c|}{$4.9_{0.8}$}  & $44.5_{0.1}$ \\ \bottomrule
\end{tabular}

}

\caption{Accuracy of \roberta-\modellarge models trained on different subsets of \qqp, when evaluated on the dataset's \hard test split and on \paws. Model performance is reported per label (Duplicate: \emph{D}, Not duplicate: \emph{$\neg$D}) and over all examples (\emph{Mean}). \Easy and \hard splits are created with the \minority method.
}
\label{tab:paws}
\end{table}

\paragraph{\Hard test sets are as challenging as manual challenge sets}
We further compare model performance on our \hard test splits to performance on challenge sets collected manually. Particularly, we compare the splits created with the \minority method for \mnli and \qqp, to the \hans~\cite{mccoy-etal-2019-right}  and \paws~\cite{zhang-etal-2019-paws} challenge sets, respectively. 

Our results (\tabref{hans} for \hans, \tabref{paws} for \paws) show that, when training on the full dataset, our automatically curated test splits are more difficult than the \hans challenge set, but not as challenging as \paws (Mean column). Interestingly, training on \easy splits (final row) makes the challenge sets dramatically more difficult, but our \hard splits are even more challenging in this setup---the model performs 0.9\% worse on \mnli compared to \hans, and 13.2\% worse on \qqp compared to \paws. 

We further find that \hard test splits are more \emph{diverse} than the challenge sets, as difficult instances affected by biases arise in \emph{all} labels in the \hard splits, while mostly in one label in the challenge sets. Our results suggest that bias-amplified splits can augment existing challenge sets by boosting their difficulty or uncovering instances that influence the biases they test.

\paragraph{Discussion}
Overall, bias-amplified splits prove to be extremely difficult for strong models. Such splits could be used to identify models that successfully generalize beyond substantial biases, and are more likely to overcome subtler ones.
Importantly, bias amplification remains challenging even when applied to recent datasets that contain fewer \easy instances (e.g., \anli and \wanli), or when compared to hand-crafted challenge sets. They could therefore be used to complement model evaluation on future, more challenging datasets. Finally, our splits can be created automatically for any existing dataset, even those for which model performance on the standard splits exceeds human performance, such as \mnli and \qqp.

\section{How Many \Hard Examples are Needed for Generalization?}
\label{sec:reinsert}

\begin{figure}[t]
\centering
\includegraphics[width=\columnwidth]{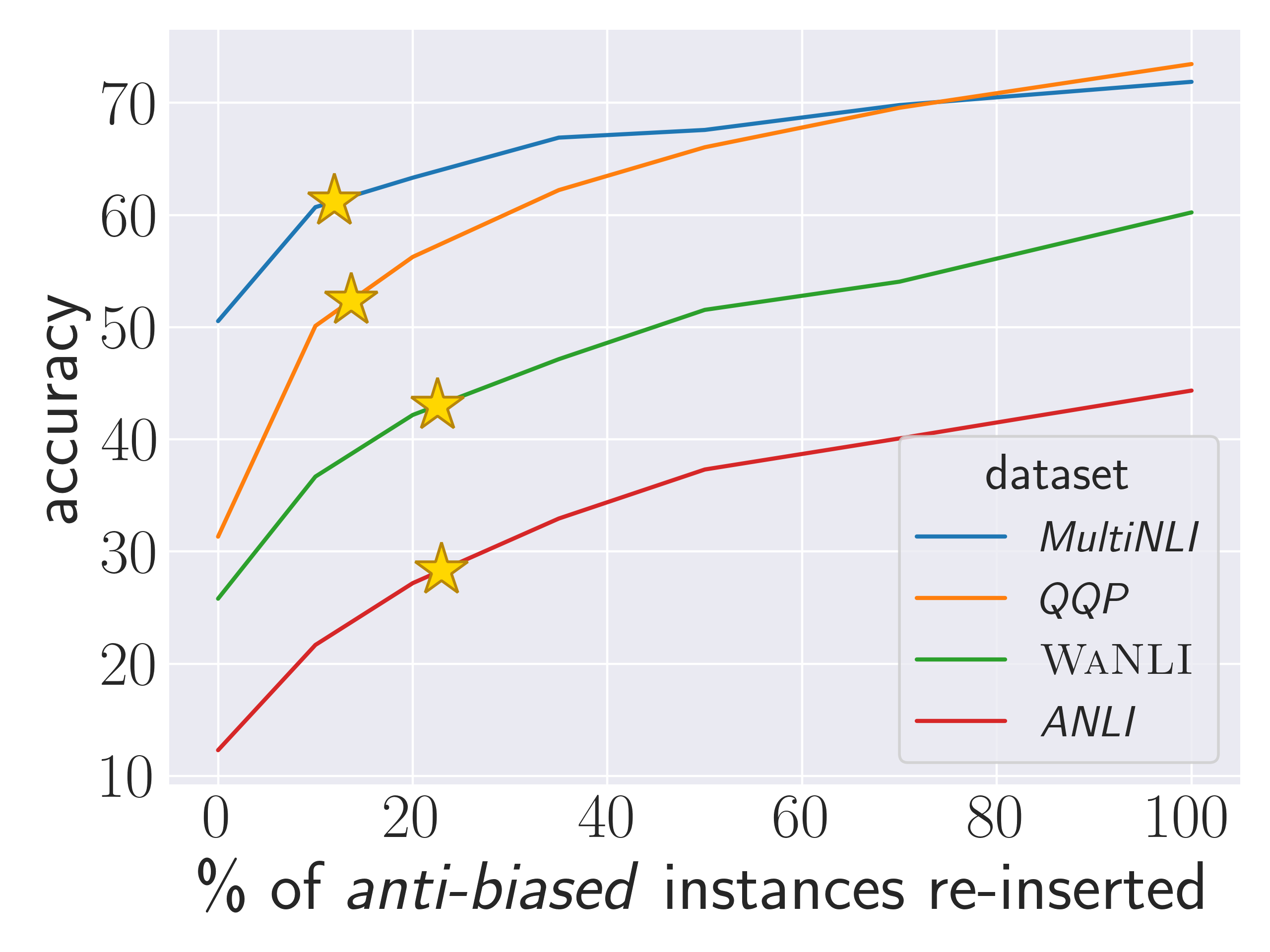}
\caption{ 
Accuracy for \roberta-\modellarge models fine-tuned on bias-amplified splits created with the \minority method, while gradually reinserting \hard instances back into the training set. Reported values are averaged across three random seeds. We interpolate and place stars (\textcolor{mustard}{$\bigstar$}) at points where the model regains 50\% of its original performance.
Models generalize from small amounts of \hard instances, but require much larger quantities to achieve comparable performance gains.
}
\label{fig:reinsertion}
\end{figure}
 
\begin{figure}[t]
\centering
\includegraphics[width=\columnwidth]{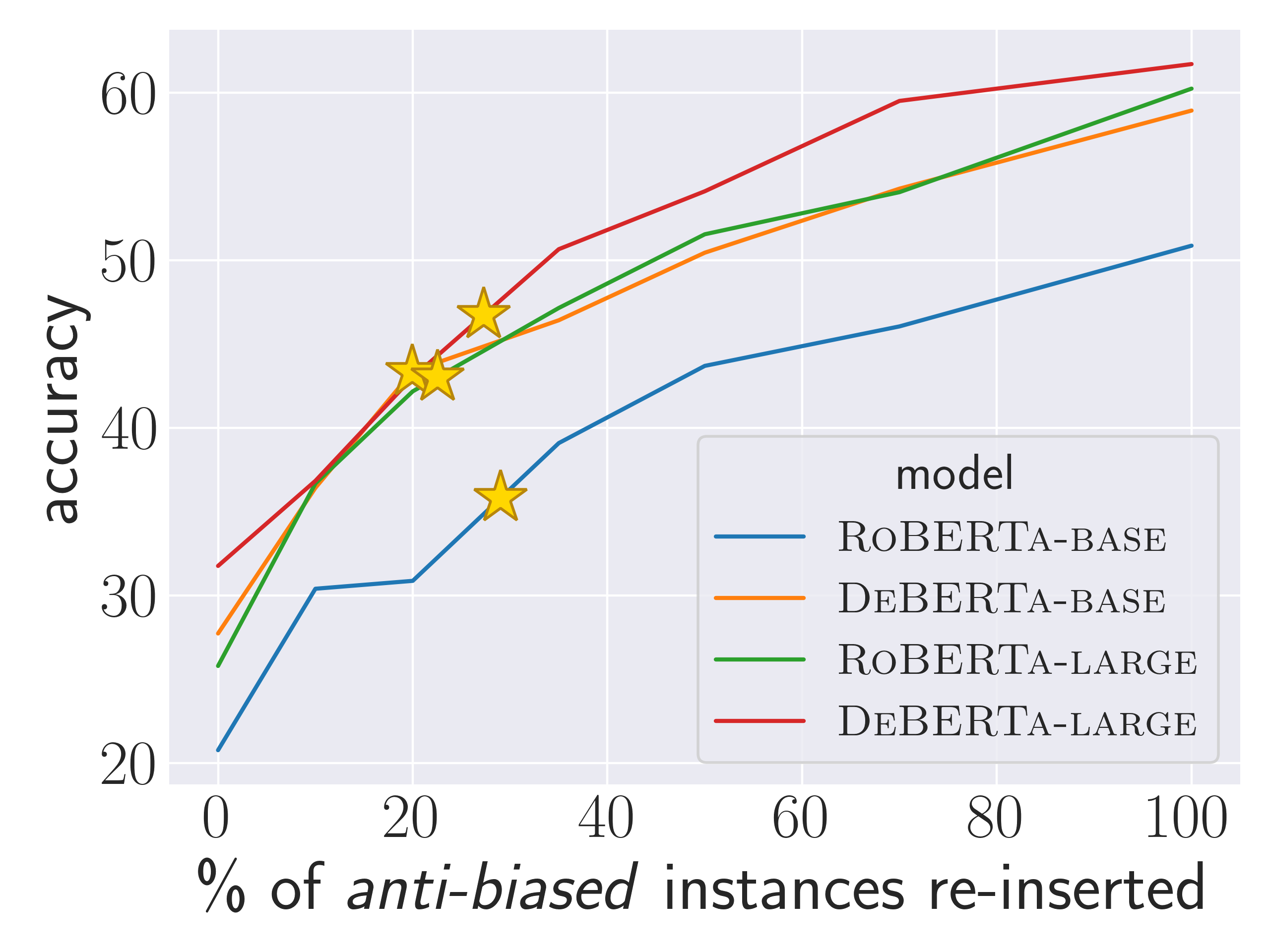}
\caption{ 
Accuracy for models trained on bias-amplified splits of \wanli created with the \minority method, while gradually reinserting \hard instances back into the training set.
}
\label{fig:reinsertion_wanli}
\end{figure}

So far, we have seen that amplifying dataset biases by eliminating \textbf{all} \hard instances from the training set uncovers shortcomings in model generalization.
We next study the effect of allowing \textbf{some} \hard instances in the training set~\cite{liu-etal-2019-inoculation}.
We fine-tune \roberta-\modellarge on all four datasets using the \easy splits created using the \minority method, while gradually reinserting 10\%, 20\%, 35\%, 50\% and 70\% of the \hard instances back into the training set.\footnote{Note that the \hard instances still constitute a minority within each cluster, as even 100\% of the \hard instances is considered a minority.}

Our results (\figref{reinsertion}) show that reinserting 20\% of the \hard training instances allows the model to close approximately 50\% of the gap from its baseline performance on the \hard test set. Surprisingly, performance grows slowly when restoring additional \hard instances, and does not match the full training set's levels even when adding 70\% of \hard instances.
This indicates that the model is capable of generalizing from small amounts of \hard instances, but is inefficient in gaining further improvements.
Results for the other models (\figref{reinsertion_wanli}) show a similar trend.

On the one hand, our results encourage careful data collection in order to fill gaps in dataset coverage 
\cite{parrish-etal-2021-putting-linguist, liu-etal-2019-inoculation}.
On the other hand, our findings indicate that data curation is not a sufficient solution, as models struggle on minority examples even when observing all available instances, and collecting more instances results in smaller further gains. Thus, it is also necessary to develop robust models that can better generalize from biased data. Our proposed framework provides a testbed for doing so.

\section{The Effect of Debiasing Methods}
\label{sec:mitigating}

Recently proposed methods were shown to be effective in improving the out-of-distribution generalization of models, either by adjusting the training loss to account for biased instances (\textbf{model debiasing};~\citealt{he-etal-2019-unlearn, clark-etal-2019-dont}), or by filtering the training set to increase the proportions of different kinds of instances found to be advantageous for generalization (\textbf{data filtering};~\citealt{aflite, yaghoobzadeh-etal-2021-increasing, jtt}).
We now examine whether such methods improve the generalization of models trained on bias-amplified training sets to \hard test instances.

We consider a \roberta-\modellarge model trained on bias-amplified splits of \mnli and \qqp based on \minority. For model debiasing, we apply the self-debiasing framework suggested by~\citet{utama-etal-2020-towards}\footnote{In self-debiasing, a biased version of the model is used to detect biases. We follow~\citet{utama-etal-2020-towards} and obtain such models by training on 2000 examples and 3 epochs for \mnli, and 500 examples and 4 epochs for \qqp.} with example reweighting~\cite{schuster-etal-2019-towards} to down-weight the loss function for biased instances; for data filtering, we apply dataset cartography to identify ambiguous instances---examples for which the model’s confidence in the gold label exhibits high variability across training epochs---and train on the 33\% most ambiguous ones, as shown to benefit generalization in~\citet{swayamdipta-etal-2020-dataset}.
Importantly, we apply both methods to the bias-amplified training split (rather than the original training set) and do not train on any other instances during the debiasing or filtering procedures.

Our results (\tabref{mitigation}) show that neither debiasing nor filtering result in substantial improvements on \hard data.
This indicates that such methods are less effective when training sets lack  sufficient \hard instances, and highlights the need for methods that could improve model generalization when additional data curation is impractical.
Our findings are also in line with recent results showing that various robustness interventions struggle with improving upon standard training in real-world distribution shifts~\cite{wilds} or dataset shifts~\cite{taori2020measuring,awadalla2022exploring}.

\begin{table}[t]
% \footnotesize
\centering
\resizebox{0.98\columnwidth}{!}{%

\begin{tabular}{@{}lcc@{}}
\toprule
               & \mnli-\hard   & \qqp-\hard    \\ \midrule
\easy           & $50.5_{1.2}$ & $31.3_{0.4}$ \\ \midrule
self-debiasing & $50.7_{0.9}$ & $33.1_{1.7}$ \\
ambiguous filtering & $51.4_{0.4}$ & $32.5_{1.9}$ \\ \midrule
100\% train    & $71.9_{0.3}$ & $73.5_{0.3}$ \\ \bottomrule
\end{tabular}

}

\caption{Accuracy of \roberta-\modellarge models trained on \mnli and \qqp with different training schemes: a \easy subset, two debiasing methods applied to the \easy subset, and the full training set. We use the \easy and \hard splits created with the \minority method. 
Applying model debiasing or data filtering approaches in the bias-amplified setting results in only slight improvements on the \hard test sets.
}
\label{tab:mitigation}
\end{table}

\section{Related Work}
\label{sec:related_work}
\paragraph{Biased splits}
The concept of re-organizing a dataset's training and test splits is often used to create more challenging evaluation benchmarks from existing datasets by inserting bias into the training set. 
\citet{sogaard-etal-2021-need} showed that using biased splits better approximates real-world performance compared to standard, random splits.
\citet{wilds} and \citet{breeds} simulated real-world distribution shifts by filtering out different kinds of data from the training and test sets, based on manually crafted heuristics. \citet{dontjustassume}  ignored the dataset's original training and test splits altogether and re-split instances to create biased splits for VQA using dataset-specific heuristics. 
Unlike such approaches, our method automatically constructs biased splits using dataset-agnostic approaches, and follows the original training and test splits.
Concurrently to this work,~\citet{godbole-jia-2023-benchmarking} re-split datasets by placing all examples that are assigned lower likelihood by an LM in the test set, and more likely examples in the training set. In some sense, that work also creates an ``easy'' training set and a ``hard'' test set, and can thus be considered a special case of our approach.

\paragraph{Challenge sets}
Given the exceptional performance of modern NLP tools on standard benchmarks, challenging test sets were created to better assess model capabilities across various tasks~\citep{isabelle-etal-2017-challenge,naik-etal-2018-stress, marvin-linzen-2018-targeted}. Such approaches often rely on human experts to identify model weaknesses and create challenging test cases using instance perturbations~\cite{jia-liang-2017-adversarial, glockner-etal-2018-breaking, belinkov2018synthetic, gardner-etal-2020-evaluating} or rule-based data creation protocols~\cite{mccoy-etal-2019-right, jeretic-etal-2020-natural}. 
Some approaches automated certain parts of these procedures, yet still require human design or annotation~\citep{bitton-etal-2021-automatic, li-etal-2020-linguistically, rosenman-etal-2020-exposing}. 

Inserting instances from challenge sets to the training set was shown to potentially alleviate their difficulty~\citep{liu-etal-2019-inoculation}, perhaps similarly to how model performance in our framework improves when reintroducing \hard examples to the training set (\secref{reinsert}).
Other work extracted challenging test subsets from existing benchmarks for focused model evaluation \cite{gururangan-etal-2018-annotation}. 
Our framework can similarly be used to better evaluate model generalization, but without requiring additional annotations or task-specific expertise, and using data that was collected in the exact same procedure as the model's training data. We further showed (\secref{results}) that our framework can be used along with existing challenge sets to increase their difficulty.

\paragraph{Dataset balancing}
Recent work proposed methods to collect benchmarks with balanced and ideally unbiased training and test splits. Such benchmarks often use a model-in-the-loop during data collection and task crowd workers to write examples on which models fail~\citep{bartolo2020, nie-etal-2020-adversarial,kiela-etal-2021-dynabench, talmor2021commonsenseqa}, or used
adversarial filtering to remove examples from existing or newly collected datasets that were easily solved by models~\citep{zellers-etal-2018-swag,zellers-etal-2019-hellaswag,dua-etal-2019-drop,aflite, sakaguchi2021winogrande}.
\citet{parrish-etal-2021-putting-linguist} proposed to use an expert linguist-in-the-loop during crowdsourcing to improve data quality and diversity.
Other work used generative methods to enrich existing datasets and compose new machine-generated examples similar to challenging seed examples~\cite{lee2021neural,wanli}.
Other studies argued that despite our best efforts, we may never be able to create datasets that are truly balanced~\cite{linzen-2020-accelerate,schwartz-stanovsky-2022-limitations}.
Our framework can be used to expose biases in such datasets and to automatically augment them with more challenging evaluation splits. 

\section{Conclusion}
\label{sec:conclusion}

Recent approaches in NLP attempted to eliminate dataset biases from training sets to produce robust models and reliable evaluation settings, yet model generalization remains a challenge, and subtler biases persist.
In this work, we argued that to promote robust modeling, models should instead be evaluated on datasets with \textit{amplified} biases, such that only true generalization will result in high performance.  We presented a simple framework to automatically create bias-amplified splits for a given dataset, finding that such splits are difficult for strong models when created for either obsolete or difficult datasets, and could potentially expose differences in generalization capabilities between models.
Our results indicate that bias amplification could ease the creation of robustness evaluation tests for new datasets, as well as inform the development of robust methods.

\section*{Acknowledgments}
We thank Inbal Magar, Gal Patel, Gabriel Stanovsky and Hila Gonen for their insightful comments that contributed to this paper.
We also thank our anonymous reviewers for their constructive feedback. This work was supported in part by the Israel Science Foundation (grant no. 2045/21).

\section*{Limitations}

In our experiments, we  evaluated models by fine-tuning on bias-amplified splits, but we did not explore the robustness of few-shot methods. Such methods are intuitively less likely to be affected by slight changes in the distribution of examples they observe. However, recent work has shown that they could still be affected by dataset biases~\cite{utama-etal-2021-avoiding, li2021systematic}, and we will use our framework to explore this in future work.

We note that our approach is less suitable for datasets with relatively small test sets. In such cases, extracting an \hard test split, which consisted of 13-21\% of the original test set in the benchmarks we considered, will result in a test set too small to reliably evaluate models. However, the methods we used to extract bias-amplified splits~(\secref{methods}) could be tuned to produce larger test sets (while keeping the amount of \hard instances in the training set relatively small), e.g., by selecting a lower number of \easy training instances ($q$, \secref{exp_setting}).

Throughout this paper, we used the term ``bias'' to describe statistical regularities in datasets that can be exploited by models as unintended shortcut solutions. While we do not explore model robustness to other types of data biases (e.g., different kinds of societal biases) our framework could potentially be used to evaluate how models handle such cases by revising the definitions of \easy and \hard instances used to create the evaluation splits.
We leave such applications of our framework to future work.

\bibliography{anthology,custom}
\bibliographystyle{acl_natbib}
\clearpage

% \clearpage
\appendix

\section{Experimental Details}
\label{sec:app_experimental}

\begin{table*}[t]
% \footnotesize
\centering
\resizebox{\textwidth}{!}{%
\setlength{\tabcolsep}{3pt}
\begin{tabular}{@{}llllllllllllllllll@{}}
\toprule
& \multicolumn{1}{c}{} &
  \multicolumn{4}{c}{\mnli} &
  \multicolumn{4}{c}{\qqp} &
  \multicolumn{4}{c}{\wanli} &
  \multicolumn{4}{c}{\anli} \\
& \multicolumn{1}{c}{} &
  \multicolumn{1}{c}{Orig.} &
  \multicolumn{1}{c}{Cart.} &
  \multicolumn{1}{c}{ParIn} &
  \multicolumn{1}{c|}{Mino.} &
  \multicolumn{1}{c}{Orig.} &
  \multicolumn{1}{c}{Cart.} &
  \multicolumn{1}{c}{ParIn} &
  \multicolumn{1}{c|}{Mino.} &
  \multicolumn{1}{c}{Orig.} &
  \multicolumn{1}{c}{Cart.} &
  \multicolumn{1}{c}{ParIn} &
  \multicolumn{1}{c|}{Mino.} &
  \multicolumn{1}{c}{Orig.} &
  \multicolumn{1}{c}{Cart.} &
  \multicolumn{1}{c}{ParIn} &
  \multicolumn{1}{c}{Mino.} \\ \midrule
   \parbox[t]{2mm}{\multirow{3}{*}{\rotatebox[origin=c]{90}{\textit{Train}}}} & \textit{full} &
  $91.1_{0.1}$ &
  $64.4_{0.6}$ &
  $81.4_{0.3}$ &
  \multicolumn{1}{l|}{ $74.3_{1.0}$} &
  $93.0_{0.0}$ &
  $66.0_{0.2}$ &
  $81.3_{0.1}$ &
  \multicolumn{1}{l|}{$77.6_{0.7}$} &
  $77.1_{0.5}$ &
  $26.4_{2.9}$ &
  $62.6_{2.0}$ &
  \multicolumn{1}{l|}{$61.7_{1.3}$} &
  $67.5_{0.5}$ &
  $35.0_{1.2}$ &
  $50.0_{0.6}$ &
  $58.3_{0.6}$ \\
& \textit{rand} &
  $91.1_{0.1}$ &
 $64.7_{0.7}$ &
  $81.5_{0.4}$ &
  \multicolumn{1}{l|}{$75.1_{0.6}$} &
  $92.8_{0.1}$ &
  $65.5_{0.3}$ &
  $81.1_{0.4}$ &
  \multicolumn{1}{l|}{$77.4_{0.1}$} &
  $77.3_{0.4}$ &
  $26.4_{2.1}$ &
  $60.8_{4.0}$ &
  \multicolumn{1}{l|}{$61.0_{2.6}$} &
  $67.6_{0.4}$ &
  $34.6_{1.0}$ &
  $49.1_{1.2}$  &
  $58.5_{0.8}$ \\ \cmidrule(l){2-18}
& \textit{bias} &
  $89.4^{\bigast}_{0.6}$ &
  $56.6_{0.3}$ &
  $71.8_{0.5}$ &
  \multicolumn{1}{l|}{$57.5_{0.5}$} &
  $89.4^{\bigast}_{1.8}$ &
  $52.6_{0.4}$ &
  $63.9_{0.3}$ &
  \multicolumn{1}{l|}{$36.8_{0.9}$} &
  $76.6^{\bigast}_{0.9}$ &
  $22.2_{1.1}$ &
  $49.6_{1.0}$ &
  \multicolumn{1}{l|}{$31.8_{1.5}$} &
  $60.2^{\bigast}_{2.6}$ &
  $12.9_{0.6}$ &
  $28.3_{1.3}$ &
  $21.4_{0.8}$ \\ \bottomrule
\end{tabular}

}
\caption{Accuracy of our approach with \deberta-\modellarge models. Different rows correspond to different training schemes: the full dataset (\textit{full}), a biased subset (\textit{bias}) and a random subset the size of \textit{bias} (\textit{rand}). Column groups correspond to different datasets. Individual columns represent testing schemes: the original validation/test set (Orig.)
and the \hard test splits: \cartography (Cart.), \partialinput (ParIn) and \minority (Mino.). Reported values are averaged across three random seeds, with standard deviation as subscripts.
Results in the last row (\textit{bias}) are of training on the \easy split and testing on the respective \hard split, except for Orig.~values (marked with $\bigast$), which are averaged over runs on all three \easy splits.
} 
\label{tab:deberta_main_results}
\end{table*}

\subsection{Datasets}
\label{sec:datasets}
We experiment with four large datasets: \qqp, \mnli, \wanli and \anli. We also run a hyperparameter search on \sst, and evaluate model performance on \hans and \paws.
Sizes of the different datasets are reported in \tabref{dataset_sizes}. 
Our implementation loads all datasets from Huggingface Datasets Hub using the \textsl{datasets} python library~\cite{datasets}.  All datasets are for  English tasks.

\paragraph{\qqp}
We experiment with the Quora Question Pairs\footnote{\url{https://quoradata.quora.com/First-Quora-Dataset-Release-Question-Pairs}} (\qqp) dataset using the version released under the GLUE benchmark~\cite{wang-etal-2018-glue}. \qqp is a dataset for the task of predicting whether pairs of questions have the same intent, i.e., if they are duplicates or not. The dataset is based on actual data from Quora.

\paragraph{Natural Language Inference (NLI)}
The task of natural language inference involves predicting the relationship between a premise and hypothesis sentence pair.
The label determines whether the hypothesis entails, contradicts or is neutral to the premise.

\paragraph{\mnli}
We experiment with the multi-genre MultiNLI dataset~\cite{williams-etal-2018-broad}, which was crowdsourced by tasking annotators to write hypotheses to a given premise for each of the three labels.
MultiNLI contains ten distinct premise genres of written and spoken data: (Face-to-face, Telephone, 9/11, Travel, Letters, Oxford University Press, Slate, Verbatim, Government and Fiction, of which five are included in the train and dev-matched sets. We don't use the dev-mismatched set in our experiments.
We use the version released under the GLUE benchmark~\cite{wang-etal-2018-glue}.

\paragraph{Adversarial NLI}
We experiment with Adversarial NLI (ANLI)~\cite{nie-etal-2020-adversarial}, a large-scale human-and-model-in-the-loop natural language inference dataset collected over multiple rounds, using BERT~\cite{devlin-etal-2019-bert}
and \roberta~\cite{roberta} as adversary models. Although each of the dataset's rounds can be used as separate evaluation settings (e.g., training on the first round and testing on the second), the data collected over all rounds can also be concatenated and used for training and evaluation; both settings were used in the original paper. In our experiments we take the concatenation approach.

\paragraph{\wanli}
We experiment with \wanli ~\cite{liu-etal-2022-wanli}, an NLI dataset collected based on worker and AI collaboration. \wanli was created by identifying examples with challenging reasoning patterns in \mnli and using a LLM to compose new examples with similar patterns. The generated examples were then automatically filtered, and finally revised and labeled by human crowd-workers. \wanli is more challenging to models than \mnli, and using \wanli instances for training was shown to improve out-of-distribution generalization.

\paragraph{SST-2} We run a hyperparameter search on \sst.
The Stanford Sentiment Treebank~\cite{socher-etal-2013-recursive} is a sentiment analysis corpus with fully labeled parse trees for single sentences extracted from movie reviews.
\sst refers to a binary classification task on  sentences extracted from these parse tress (negative or somewhat negative vs somewhat positive or positive, with neutral sentences discarded). We use the version of \sst released under the GLUE benchmark~\cite{wang-etal-2018-glue}.

\paragraph{\hans}
We evaluate models on \hans (Heuristic Analysis for NLI Systems; \citealt{mccoy-etal-2019-right}), a challenge set used to assess whether NLI models adopt invalid syntactic heuristics that succeed for the majority of NLI training examples (e.g., lexical overlap implies that the label is entailment), instead of learning more generalizable solutions. \hans contains many \emph{entailment} examples that support these heuristics, and many \emph{non-entailment} examples where such heuristics fail. When evaluating NLI models that were trained with 3-way labels (as in \mnli), we map \emph{contradiction} or \emph{neutral} predictions to the \emph{non-entailment} label. \hans was created by automatically filling in words in templates devised by human experts.

\paragraph{\paws}
We evaluate models on \paws~(Paraphrase Adversaries from Word Scrambling; \citealt{zhang-etal-2019-paws}), a challenge set for the paraphrase identification task that focuses on non-paraphrase pairs with high lexical overlap. Challenging pairs are generated by controlled word swapping and back translation, followed by fluency and paraphrase judgments by human raters. We evaluate models on the test set of the $\textsl{PAWS}_{\text{Wiki}}$ dataset.

\begin{table}[t]
\centering

\begin{tabular}{@{}lccc@{}}
\toprule
       & Train   & Validation  & Test  \\ \midrule
\qqp  & 363,846 & 40,430 & -     \\
\mnli  & 392,702 & 9,815 & -     \\
\anli  & 162,865 & 3,200 & 3,200 \\
\wanli & 102,885   & 5,000 & -     \\
\sst & 67,349   & 872 & -     \\ 
\hans & -   & - & 30,000     \\ 
\paws & -   & - & 8,000     \\  \bottomrule
\end{tabular}%

\caption{Datasets sizes. Development set in MultiNLI is the matched validation set (we did not use the mismatched validation set).}
\label{tab:dataset_sizes}
\end{table}
\begin{table*}[t]
\centering

\begin{tabular}{@{}lcccccccc@{}}
\toprule
 & \multicolumn{2}{c}{\mnli} & \multicolumn{2}{c}{\qqp} & \multicolumn{2}{c}{\wanli} & \multicolumn{2}{c}{\anli} \\
              & train   & test  & train   & test  & train  & test & train   & test \\ \midrule
\cartography  & 309,873 & 2,070 & 297,735 & 7,346 & 89,402 & 656  & 134,068 & 566  \\
\partialinput & 309,873 & 2,070 & 297,735 & 7,346 & 89,402 & 656  & 134,068 & 566  \\
\minority     & 309,873 & 2,044 & 297,735 & 7,462 & 89,402 & 637  & 134,068 & 938  \\ \bottomrule
\end{tabular}

\caption{Sizes of the train and test bias-amplified splits created with each of the considered methods (\secref{methods}). 
Since the number of \easy train instances is induced by the clustering in the \textsl{minority examples} approach, but is a hyperparameter $q$ for the two other approaches, we simply adjust $q$ to create equally sized training sets for all three methods. 
We use the same $q$ used for choosing \easy \textbf{train} instances when choosing \hard \textbf{test} instances. We note that for the \minority method, the training set clustering and the predicted test set clustering (based on a simple nearest neighbor classifier fitted on the training set) are two \emph{different} clusterings, which can result in different proportions of \minority between the train and test sets. This explains the difference in the amounts of \hard test instances between \minority and the other two methods.}
\label{tab:split_sizes}
\end{table*}

\subsection{Experimental Settings}
\label{sec:settings}

We experiment with the \base and \modellarge variants of \roberta~\cite{roberta} and \deberta~\cite{deberta}. 
Our implementation and pretrained model checkpoints use the Huggingface Transformers library~\cite{wolf-etal-2020-transformers}. For \deberta, we use the latest v3 checkpoints.
When partitioning datasets to \easy and \hard sub-parts, we use the training dynamics and representations of a \roberta-\base model. We create \easy training and \hard test sets based on  a single run of the model. All further experiments (e.g., training \deberta-\modellarge on \easy instances and testing it on \hard instances) are run with 3 random seeds, using the same train and test splits. 

\paragraph{Bias-amplified split sizes}
\tabref{split_sizes} reports the sizes of the bias-amplified \easy train and \hard test splits created based on each of the three methods (\secref{methods}) we experimented with.

\begin{table}[t]
\centering
\resizebox{\columnwidth}{!}{%
\footnotesize
\begin{tabular}{@{}lcccc@{}}
\toprule
Hyper-parameter     & \base & \modellarge \\ \midrule
Warmup Ratio            & 0.06 & 0.06   \\
Learning Rate           & 1e-5 & 1e-5  \\
Learning Rate Decay & Linear         & Linear             \\
Batch Size              & 32   & 32    \\
Max. Train Epochs       & 10   & 5      \\
Early Stopping Patience & 3    & 3   \\ \bottomrule
\end{tabular}
}
\caption{Hyperparamets for finetuning \roberta.}
\label{tab:finetune_hyperparams_roberta}
\end{table}
\begin{table}[t]
\centering
\resizebox{\columnwidth}{!}{%
\begin{tabular}{@{}lcccc@{}}
\toprule
Hyper-parameter     & \base & \modellarge \\ \midrule
Warmup Steps      & 100    & 100  \\
Learning Rate         & 1.5e-5 & 1e-5 \\
Learning Rate Decay & Linear         & Linear                   \\
Batch Size              & 32     & 32   \\
Max. Train Epochs        & 10     & 5    \\
Early Stopping Patience   & 3      & 3    \\ \bottomrule
\end{tabular}
}
\caption{Hyperparamets for finetuning \deberta.}
\label{tab:finetune_hyperparams_deberta}
\end{table}
\paragraph{Hyperparameters} For fine-tuning, we did not optimize the hyperparameters and instead used parameters that were included in the hyperparameter search on down-stream tasks from the original papers, except for training \modellarge models for 5 epochs instead of 10.  We also used an early-stopping patience threshold of 3 epochs. 
We report all fine-tuning hyperparameters in \tabref{finetune_hyperparams_roberta} and \tabref{finetune_hyperparams_deberta}.
% We used a maximum sequence length of 128 tokens with longer sequence being truncated, starting from the longest sentence of the sentence-pair input.

\begin{table}[t]
\centering
\resizebox{\columnwidth}{!}{%
\begin{tabular}{@{}lcccc@{}}
\toprule
Datasets & \multicolumn{2}{c}{\roberta} & \multicolumn{2}{c}{\deberta} \\
         & \base      & \modellarge     & \base      & \modellarge     \\ \midrule
\mnli    & 8          & 8               & -          & 10              \\
\qqp     & 8          & 8               & -          & 10              \\
\wanli   & 2          & 4               & 2          & 4               \\
\anli    & 4          & 5               & -          & 5               \\
\sst     & 1          & -               & -          & -               \\ \bottomrule
\end{tabular}
}
\caption{Average runtimes for fine-tuning, in hours.}
\label{tab:finetune_runtimes}
\end{table}
\paragraph{Average runtimes}
For \roberta-\base, each train run was performed on a single RTX 2080Ti GPU (10GB). For all other models, each train run was performed on a single Quadro RTX 6000 GPU (24GB). 
We report average runtimes (training and inference combined) in \tabref{finetune_runtimes}.

% 

% for bias-amplified splits: sizes of hard test subsets for each method, easy train subsets

\section{Additional Results}
\label{sec:app_results}
\subsection{Main Results for \deberta}
\label{deberta_main_results}
\tabref{deberta_main_results} shows our results for \deberta-\modellarge for the experiment described in \secref{exp_setting}.

\section{Clustering Algorithm for Detecting Minority Examples}
\label{sec:app_clustering}
Minority examples~\cite{tu-etal-2020-empirical, sagawa2020investigation} are often detected by searching for spurious features correlated with one label in the instances of another label (e.g., high word overlap between two non-paraphrase texts).
Motivated by recent work that leverages [CLS] token similarity in fine-tuned models between different instances~\cite{liu-etal-2022-wanli, pezeshkpour-etal-2022-combining}, we proposed a  model-based clustering approach to automatically detect minority examples (\secref{methods})

Our approach is based on simple analyses applied to the clustering of a given dataset's [CLS] model representations. In this work we used the deep clustering algorithm described in \secref{methods}, \deepcluster~\cite{deepcluster}, to perform the clustering. In this appendix we provide more details on the algorithm (App. \ref{subsec:deepcluster}), its implementation details (App. \ref{subsec:clustering_implementation}), and the hyperparameter search we ran to select a good configuration (App. \ref{subsec:clustering_hyperparameters}). We also show preliminary results for using alternative clustering methods for detecting \minority (\ref{clustering_standard}) and for the difficulty of the bias-amplified splits based on \minority detected over different random seeds (App. \ref{sssec:clustering_over_seeds}).

\subsection{\deepcluster}
\label{subsec:deepcluster}

\textsc{DeepCluster} alternates between grouping the model's representations with a standard clustering algorithm to produce pseudo-labels, and updating the parameters of the model by predicting these pseudo-labels.
To apply \deepcluster to BERT-like models,\footnote{\deepcluster is used in the original paper for pretraining Computer Vision models.} we consider a model fine-tuned on the dataset to be clustered.\footnote{Importantly, the model is fine-tuned on the \textit{task label}, rather than the clustering label (such labels do not exist a priori, but are generated automatically).} We extract and cluster its [CLS] token representations using a standard clustering algorithm, and then perform one \deepcluster iteration by fine-tuning a new pretrained model with the pseudo-labels (instead of the dataset's gold labels) for one epoch.\footnote{We emphasize that this pretrained was never fine-tuned on any task labels, but only on the pseudo-labels.} We then cluster the representations from this second model to obtain the final clustering.

\subsection{Implementation Details}
\label{subsec:clustering_implementation}
As the standard clustering algorithm at the base of \deepcluster, we use Ward's method~\cite{ward1963}, a popular hierarchical clustering algorithm which is deterministic and therefore stable across different runs, a quality which we found preferable. 
We use the fastcluster~\cite{fastcluster} python implementation with the default settings.  

\paragraph{Applying Ward's clustering to large-scale datasets}
We did not have resources with enough memory to cluster the entire training sets of \mnli and \qqp, which contain more than 320k examples. We therefore approximate the clustering assignment by clustering a random sample of 50\% of the training set, and then using a simple nearest-neighbor classifier to predict the assignments for the other 50\%.\footnote{We fit the classifier on the clustered sample's representations as inputs and clustering assignments as output labels.}

\paragraph{Runtime}
Running \deepcluster requires (1) fine-tuning a model for 1 epoch and then extracting its representations, which takes 15--70 minutes on a GPU, and (2) clustering the representations on a CPU, which takes 40 minutes for \wanli and \anli, and 3 hours for \mnli and \qqp.

\subsection{\deepcluster Hyperparameters}
\label{subsec:clustering_hyperparameters}
\deepcluster has three hyperparameters: the number of final clusters $k$, the number of pseudo-labels $m$ for representation learning, and the Transformer layer from which [CLS] representations are extracted for clustering. We used $k=10$ clusters for all datasets, and searched for a good configuration for the other two hyperparameters on \sst~\cite{socher-etal-2013-recursive}, which were then used for experiments on all other datasets in the paper. 
We searched over $m\in\{10,30,50,100,300,500,1000,1500,3000\}$\footnote{The number of pseudo-labels $m$ can differ from the desired number of final clusters $k$. Setting $m\gg k$ yielded better results in the original paper.} and representations from the last four layers of \roberta-\base.

For each set of hyperparameters, we applied the \minority method to create \easy training and \hard test splits, and trained two \roberta-\base models---one on the \easy train split, and a baseline model on an equally-sized random train subset, finally choosing the hyperparameters that lead to the largest performance drop on \hard test instances between the two. 
The best hyperparameters were the layer before last of \roberta-\base (layer 11) and $m=1500$.

\begin{figure}[t]
\centering
\includegraphics[width=\columnwidth]{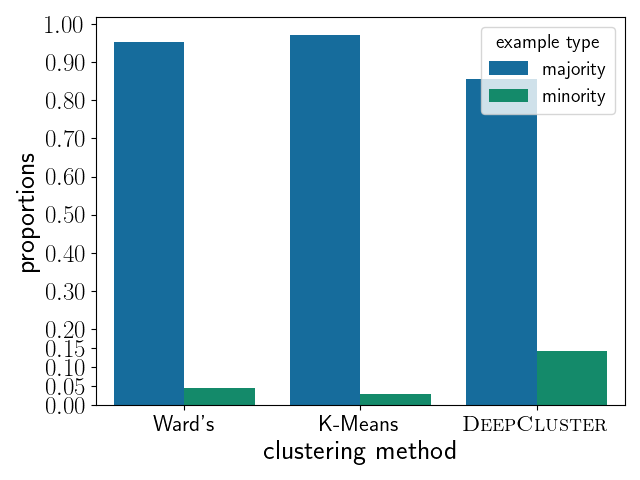}
\caption{ 
The mean proportions of majority and minority label instances within clusters for different clusterings of \sst, based on the [CLS] representations of \roberta-\base fine-tuned on the dataset. [CLS] tokens are taken from the layer before last of the model.
}
\label{fig:standard_clustering}
\end{figure}

\subsection{Preliminary Results}
\label{subsec:clustering_preliminary_results}

\subsubsection{Using Standard Clustering to Detect Minority Examples}
\label{clustering_standard}
Our preliminary experiments show that standard clustering algorithms applied to the [CLS] representations of models fine-tuned on the original task tend to create label-homogeneous clusters, i.e., they are less likely to cluster together instances from different labels. In \figref{standard_clustering} we show the average proportions of majority and minority instances within clusters for different clusterings of \sst (which has two task labels) based on \roberta-\base representations. We compare \deepcluster and two standard clustering algorithms: K-Means and Ward's method. We find that the clusters of standard methods contain, on average, less than 5\% minority label instances, while clusters based on \deepcluster are more label-diverse and contain 15\% minority label instances. When inspecting how many individual clusters contain more than 10\% minority label instances, we find that for both standard methods \textbf{only one cluster} (out of 10) meets this threshold, whereas there are \textbf{6 such clusters} with \deepcluster.

\subsubsection{Difficulty of Minority Examples in Bias-amplification Over Random Seeds}
\label{sssec:clustering_over_seeds}
We ran a preliminary experiment on \sst to examine whether the difficulty of the bias-amplified splits based on the \minority method varies with the seed used to collect data representations. We clustered \sst using \deepcluster based on representations of \roberta-\base. We used 3 different seeds to fine-tune the model and run \deepcluster, and created a bias-amplified split from each resulting clustering. We then examined the performance drops between a \roberta-\base model trained on the \easy vs. random split (as in the hyperparameter search; see App. \ref{subsec:clustering_hyperparameters}). 
The mean absolute performance drop was -16.7, with a standard deviation of 5.9. This indicates that while there is variation between seeds, all clusterings produced challenging settings. We conclude that when seeking to create the \emph{most} challenging splits, running a hyperparameter search over multiple seeds on the dataset the splits are created for would likely lead to better results. In this work, we did not optimize the clustering hyperparameters for each dataset, and therefore used one seed for all clusterings.
% for clustering: need some results on k-means, ward's. need 

% \input{appendix/A4.analysis_settings}
% \input{appendix/A5.clustering}
% \input{appendix/A6.detailed_results}

\end{document}